\documentclass[conference]{IEEEtran}
\IEEEoverridecommandlockouts

\usepackage{cite}
\usepackage{amsmath,amssymb,amsfonts}
\usepackage{algorithmic}
\usepackage{graphicx}
\usepackage{textcomp}
\usepackage{xcolor}
\def\BibTeX{{\rm B\kern-.05em{\sc i\kern-.025em b}\kern-.08em
    T\kern-.1667em\lower.7ex\hbox{E}\kern-.125emX}}
\usepackage{hyperref}

\usepackage{mathptmx} 
    
\begin{document}

\title{Continuously Learning New Words\\
in Automatic Speech Recognition}

\author{\IEEEauthorblockN{Christian Huber$^1$}
\and
\IEEEauthorblockN{Alexander Waibel$^{1,2}$}
}

\maketitle

\begin{abstract}
Despite recent advances, Automatic Speech Recognition (ASR) systems are still far from perfect. Typical errors include acronyms, named entities, and domain-specific special words for which little or no labeled data is available. To address the problem of recognizing these words, we propose a self-supervised continual learning approach: Given the audio of a lecture talk with the corresponding slides, we bias the model towards decoding new words from the slides by using a memory-enhanced ASR model from the literature. Then, we perform inference on the talk, collecting utterances that contain detected new words into an adaptation data set. Continual learning is then performed by training adaptation weights added to the model on this data set. The whole procedure is iterated for many talks. We show that with this approach, we obtain increasing performance on the new words when they occur more frequently (\textgreater80\,\% recall) while preserving the general performance of the model. 
\end{abstract}


\section{Introduction}

In the last decade, ASR systems improved tremendously in terms of word error rate (WER) due to more data, more computing power, and better architectures \cite{vaswani2017attention, pham2019very}.
However, these systems are still far from perfect. 
Although in principal end-to-end systems are open-vocabulary systems, when using appropriate modeling units, such as byte-pair encoded (BPE) characters, in practice, words not seen during training are often not reliably recognized.
Typical errors
are observed in the categories of
cross-lingual words (e.g. 'upgeloaded' is mixing German and English), numbers (e.g. 1945 vs. 19:45 vs. \$19.45),
acronyms (e.g. ICASSP),
named entities
and domain-specific special words (as they occur in specialized meetings or lectures). The word error rate (WER) is only slightly affected by these errors because they are rather infrequent. However, these words are important for understanding the content as they carry a lot of information that is lost when they are misrecognized.
Furthermore, proper interpretation of these words is critical for downstream processing such as in speech translation \cite{fugen2007simultaneous, waibel2012simultaneous}.
To measure this, we must not only evaluate WER but also recall, precision, and F1 score for these words, i.e., how often are they recognized if they occur and how often do they produce false positives.

\begin{figure}[t]
    \centering
    \vskip -10pt
    \includegraphics[page=2,trim={8.5cm 6.9cm 12.0cm 4.8cm},clip,width=1.0\columnwidth]{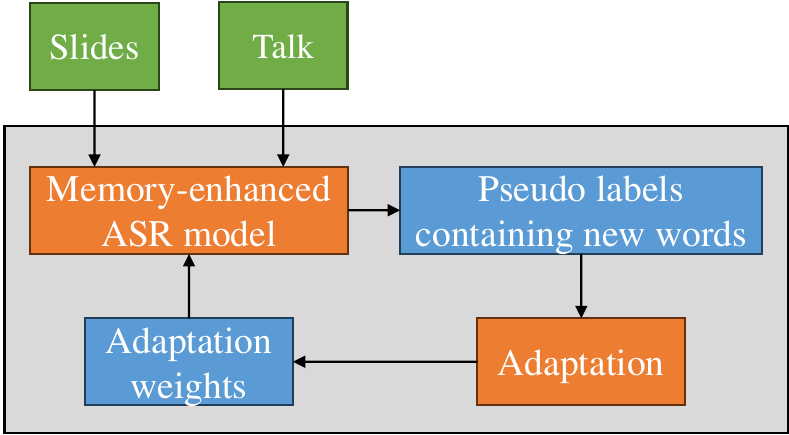}
    \vskip -5pt
    \caption{Illustration of the continual learning: In each learning cycle the model is biased towards the new words from the slides of the current talk, inference is performed and pseudo-labels containing the new words are collected; then the model is adapted.}
    \label{fig:overview}
    \vskip -10pt
\end{figure}

In this work, we tackle the problem of learning such acronyms, named entities, and domain-specific special words from scarce data in a self-supervised manner.
For this, we
\textbf{1)} show that we can adapt an ASR model to detect new words with little given labeled data using a factorization-based approach,
and \textbf{2)} use this factorization-based approach in combination with a memory-enhanced ASR model and slides of lecture talks to perform self-supervised continual learning (see Figure \ref{fig:overview} and Section \ref{sec:approach}).
We empirically show that this approach does not lead to catastrophic forgetting even for a large number of learning cycles (66), while improving the recall of new words to more than 80\% as new words occur more frequently.


\section{Related Work}

An approach to new-word learning of ASR systems
is context biasing \cite{huber2021instant,le2021contextualized,maergner2012unsupervised,suhm1993detection,suhm1994towards,waibel2015enhanced}.
The model is biased with a list of words. Depending on the approach, this list is used in different ways.
The problem is that the scale of such approaches for continual learning is limited. 



Other approaches study the
continual learning of ASR systems \cite{chang2021towards,fu2021incremental,yang2022online,chennupati2022ilasr,vander2022continual}. The access to old data is restricted and multiple regularization-based and data-based methods \cite{wang2024comprehensive} are used. The problem is that many hours of labeled adaptation data are used for each learning cycle, without considering where to get such data, and/or only a few learning cycles are considered.







As stated, many works in continual learning study the learning of new tasks while limiting the access to the old data. However, access to old data is not always a problem in production,
e.g., when publicly available data sets are used as in our case, but the amount of computation power and particularly the availability of new (labeled) data.
It is not feasible to train a new model from scratch each time a new word has to be learned.
The same holds for collecting hundreds of labeled utterances for each new word, because of cost and privacy issues. Therefore, we study how we can adapt an ASR model to detect new words with modest computation power and little labeled data or data collected in a self-supervised manner.

\section{Experiments}
\label{sec:experiments}

The experiments consist of two parts: First, we look at how to learn new words from little given labeled data.
For this, 
we experimented with a factorization approach \cite{pham2021efficient,hu2021lora}: 
To each weight $W\in\mathbb{R}^{n\times m}$ of a linear layer of the neural network, a low-rank matrix
\begin{equation*}
    \overline{W} = \sum_{i=1}^kr_is_i^T
\end{equation*}
is added, where $k\in\mathbb{N}$,
$r_i\in\mathbb{R}^n$, $s_i\in\mathbb{R}^m$, ${i\in\{1,\ldots,k\}}$.
By changing only the weights $\overline{W}$, one can efficiently adapt the model. The hyperparameter $k$, typically $k \ll \min\{n,m\}$, controls the amount of flexibility of the model, and $r_i$ and $s_i$ are denoted as factorization weights.

Second, we perform a continual learning experiment (see Figure \ref{fig:overview} and the second part of Section \ref{sec:approach}). We use the factorization approach mentioned above together with a memory-enhanced ASR model similar to \cite{huber2021instant} to first extract pseudo-labels of utterances containing new words and then adapt the model with this new data. 
The memory-enhanced ASR model consists of a memory
\begin{equation*}
Z=(Z_1,\ldots,Z_L),\qquad L\in\mathbb{N},
\end{equation*}
where each memory entry $Z_l$, $l\in\{1,\ldots,L\}$, is (the tokenized representation of) a word or a short phrase. The model extracts information from $Z$ through attention mechanisms, and therefore the prediction of the next token is biased towards the words/phrases in the memory $Z$.

\subsection{Data}

For the experiments
we created two data sets. First, we created a
\href{https://github.com/chuber11/earnings21-new_words-dataset}{new-words data set}
for the factorization approach experiments
based on \cite{del2021earnings}. We extracted words from the categories named entity (of persons), acronym (abbreviations), and domain-specific special word (products, events, laws, locations, and organizations) from the annotations. Then, we manually filtered those that occurred two or more times.
From that, we created a development split (15\,\% of the data, for hyperparameter optimization) and an evaluation split. Both contain a training set (for model adaptation) and a test set (for model evaluation). For each new word, one occurrence is used for the test set and the remaining ones for the training set.
For most new words, we have only one training example, for a few examples more.

Second, we downloaded videos of
\href{https://github.com/chuber11/cl_learning_datasets}{computer science lectures}
with corresponding slides for the continual learning experiment from YouTube. The data set collected consists of
of 66 lecture talks. From the slides (given as PDF file) we extracted words that were not in our training data and manually went over the output to filter for obvious errors, e.g., some words next to each other were merged by the text extraction process, but we did not modify the extracted words by content. This resulted in 3891 extracted words (2199 unique words) that were denoted new words.

To evaluate the overall performance of our models, we report the WER on the Tedlium testset (1155 utterances, 2.6 hours).

\subsection{Models and Approaches}
\label{sec:approach}

We use
the framework \href{https://github.com/quanpn90/NMTGMinor}{NMTGMinor} 
which is based on PyTorch and uses the Fairseq \cite{ott2019fairseq} pre-trained models. 
For the baseline model, similar to recent work, for example \cite{li2020multilingual}, we start with a transformer model \cite{vaswani2017attention, pham2019very}, where the
encoder is initialized with the pretrained Wav2Vec 2.0 model \cite{baevski2020wav2vec} and the
decoder is initialized with the decoder part of the pretrained mBART 50 model \cite{liu2020multilingual} (using the mBART 50 tokenizer).


We trained the baseline model on ASR data from Common voice, Europarl, Fisher, HOW2, Hub4, Librispeech, Switchboard, Tedlium, and Voxforge (5433 hours in total), denoted baseline data.
For some of the mentioned corpora, only lowercase labels without punctuation were available. Therefore, we trained an encoder-decoder model on Wikipedia text to map lowercase text without punctuation to cased text with punctuation and applied that model to the data.

When training the factorization approach, we freeze all weights of the baseline
model
and only train the added factorization weights until the validation perplexity on the development split does not decrease anymore.
To avoid catastrophic forgetting \cite{french1999catastrophic}, we train the model with a mixture of baseline data and new-words data. To emphasize that the model should learn the new words, we massively upsample utterances containing them \cite{huber2020supervised}. Only training the factorization weights significantly decreases training time compared to training a model from scratch or from a pre-trained model. The model is trained within a few hundred updates (compared to the baseline model which is trained for 60k updates).
Each factorization approach was trained in less than four hours on one A6000 GPU. On the other hand, the baseline model needed 60k updates to finish training.

The memory-enhanced ASR model \cite{huber2021instant} consists of the audio encoder and decoder of the baseline model and additionally has a memory encoder and a memory decoder (both with six layers and initialized with parts of the pre-trained mBART 50 model). Each memory entry is encoded independently by the memory encoder. Each memory decoder layer consists of the standard mBART decoder layer followed by a memory-attention layer and a memory-entry-attention layer extracting information from the encoded memory entries (see \cite{huber2021instant} for more details). During training of the memory-enhanced ASR model, only the memory encoder and memory decoder are trained, the rest of the model is frozen.



In the continual learning experiment (see Figure \ref{fig:overview}), we start with an empty data set of new-words adaptation data.
We iterate over the given talks and do for each talk the following four steps: 
\textbf{1)} We use the new words extracted from the corresponding slides as memory to bias the model toward the new words,
\textbf{2)} We perform inference on the talk using the model from the last iteration step (using the baseline model in the first iteration step),
\textbf{3)} We extract all utterances from the talk containing a word from the current memory or any past memory and add them together with the pseudo-labels created to the new words adaptation data. Using utterances containing a word from a past memory is critical for continual learning, since we want to use the collected data to increase the performance of new words even when they are no longer present in the memory. All utterances in which a new word occurs the second time are used for the new words validation data, and
\textbf{4)} We use the factorization approach applied to the new-words adaptation data as well as the baseline data to adapt the model. Especially, we train new factorization weights initialized randomly.
After the training, we use the best model according to perplexity on the new words validation data. These four steps are iterated for all given talks. We denote this approach by MEM-CL.
Note that it is not a severe problem if a new word is not correctly recognized by the memory-enhanced ASR model since an important new word most likely will occur multiple times, and therefore the memory-enhanced ASR model can also detect it later.

For comparison, we report the approach where no factorization weights are learned but instead the memory is always enriched by the new words of the current talk (denoted by MEM-ALL).
Furthermore, we vary this approach by keeping words in memory only if they are recognized in the talk in which they are added to the memory (denoted by \mbox{MEM-FOUND}). This is similar to MEM-CL, where only utterances containing found new words are transferred to the new-words adaptation data.

\section{Results}

\begin{figure}[t]
  \centering
  \includegraphics[trim={0.4cm, 0.2cm, 0.4cm, 0.4cm},clip,width=1.0\columnwidth]{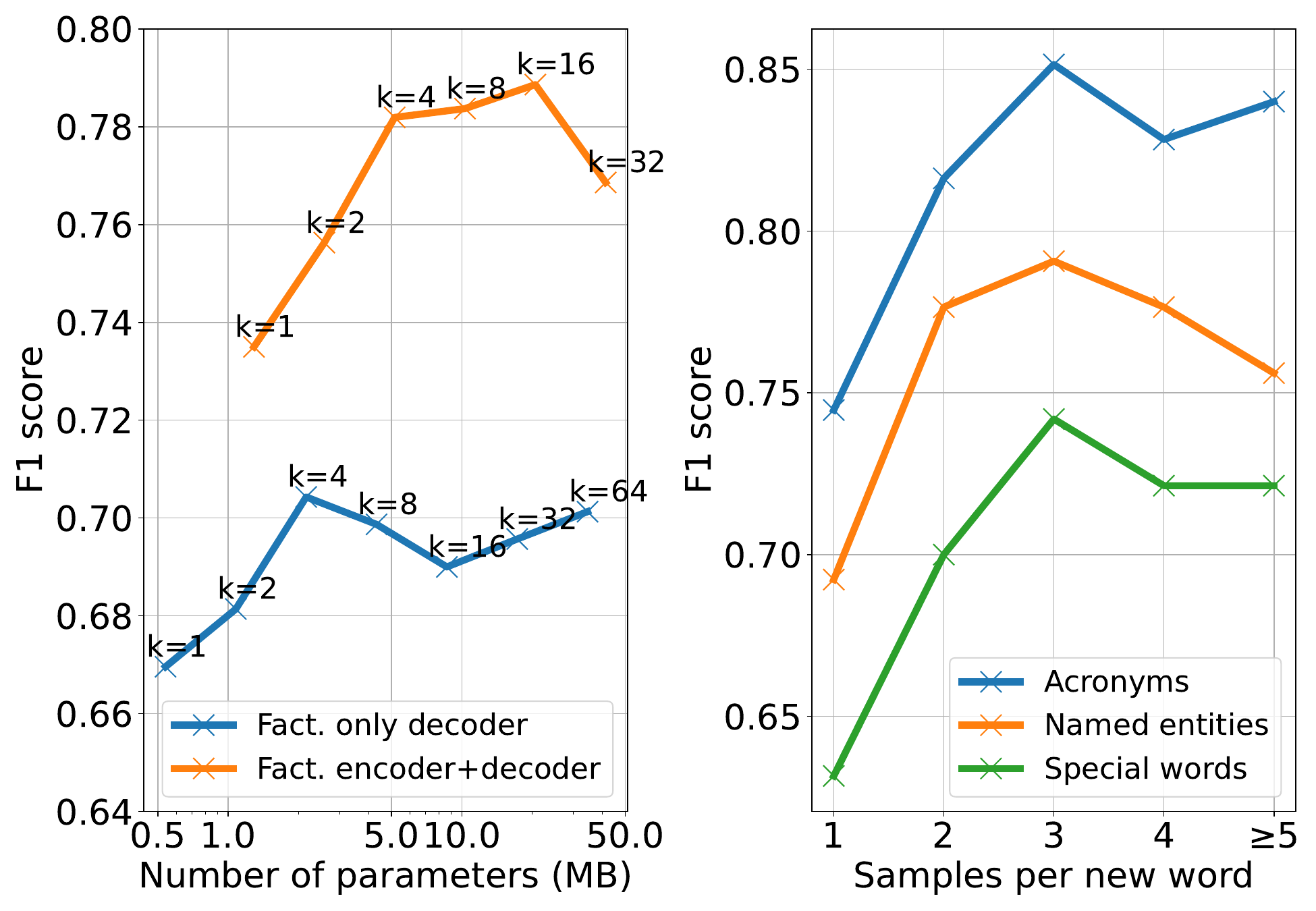}
  \vskip -5pt
  \caption{Results of the factorization experiment: Left: Number of parameters (MB, 16-bit) vs. F1-score after training with the new-words data for a factorized decoder and a factorized encoder+decoder. The baseline model has F1-score $0.402$.
  Right: Number of training samples per new word vs. F1-score for the different categories
  with a factorized encoder+decoder and $k=4$.
  }
  \label{fig:newwords}
  \vskip -10pt
\end{figure}

For the factorization experiment, we tuned the learning rate and the upsampling factor of the new-words data on the development split of the new-words data set.
In Figure \ref{fig:newwords} the results can be seen for the learning rate $10^{-4}$ and the upsampling factor $10^5$ (this corresponds to approximately half new and baseline data in each batch). We compare the addition of factorization weights to the whole model and to only the decoder.
We report only the F1 score (evaluated only on the new words) because the precision of all approaches is very high ($>\hskip -3.1pt98\,\%$) and therefore the recall strongly correlates with the F1 score.

On the left of Figure \ref{fig:newwords}, we see the amount of factorization parameters versus the F1 score evaluated on the new words. Until the point $k=4$, the performance increases for both the factorized decoder and the factorized encoder and decoder, and the model is able to use the additional flexibility.
Generally, the model with factorized encoder and decoder is better than the model with only factorized decoder. However, adapting only the factorized decoder is faster by approximately a factor of two to five.
For comparison, the memory-enhanced ASR model trained for the continual learning experiment scores recall 0.721, precision 0.964 and F1-score 0.825 when using a memory containing all new words from the eval split.
On the right of Figure \ref{fig:newwords}, the number of samples per new word versus the F1 score is shown for each category. We see a similar behavior for all categories and that the performance increases when adapting with more samples per new word. For four and five samples per new word, the performance does not increase anymore, probably since there are not many new words for which that number of samples is available. 
Furthermore, acronyms seem to work best, possibly due to the small number of characters each part of the acronym can be, followed by named entities and special words.

\begin{figure*}[t]
  \centering
  \includegraphics[trim={0.4cm 0.3cm 0.6cm 0.4cm},clip,width=1.0\textwidth]{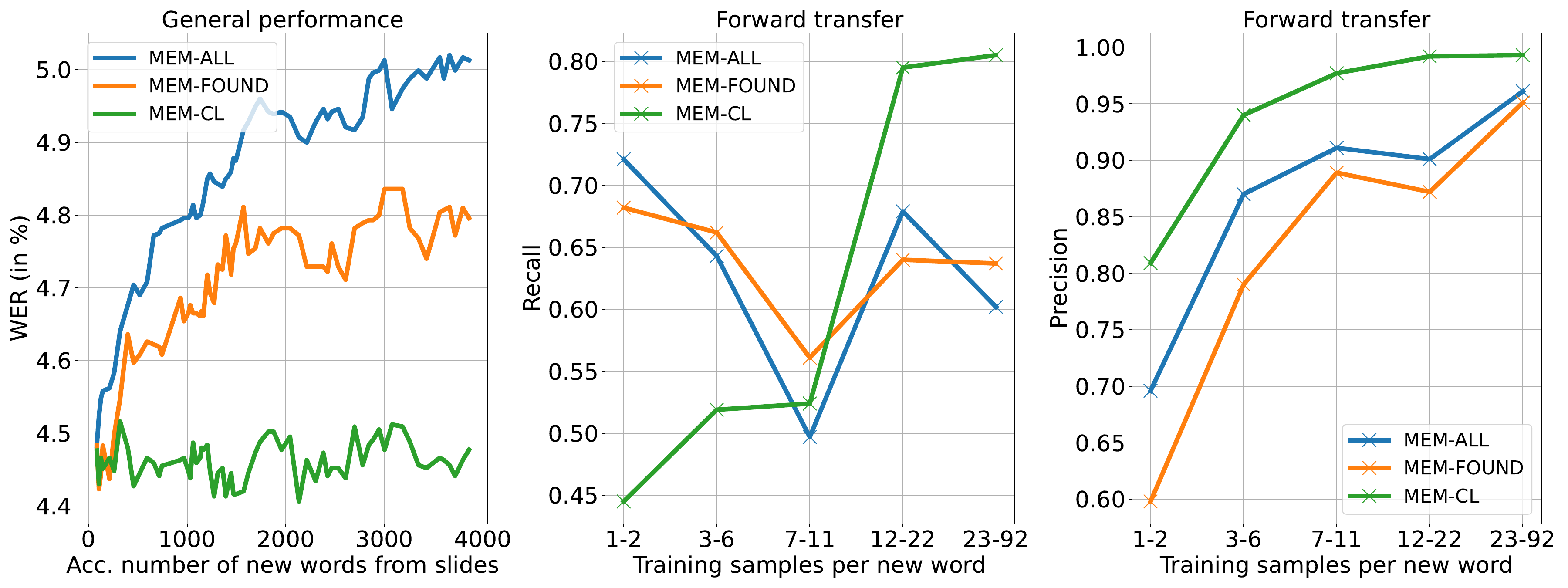}
  \vskip -5pt
  \caption{Results of the continual learning experiement: Left: General performance: Accumulated number of new words from slides versus WER (in \%) on the Tedlium testset. Middle and right: Forward transfer: Number of training samples per new word versus forward transfer recall and precision.
  }
  \label{fig:cl}
  \vskip -10pt
\end{figure*}






For the continual learning experiment, we use $k=4$, factorized decoder only to reduce the amount of computational power required and the upsampling factor $10^5$. We report the WER on the Tedlium testset over the course of the learning cycles (see Figure \ref{fig:cl}, left) as well as the forward transfer, that is, if a new word occurred, how likely is it recognized
in a later talk (see Figure \ref{fig:cl}, middle and right).
Since the baseline decoder and the memory decoder independently predict the next token (afterward the results are combined), the factorization adaptation of the baseline decoder does not interfere with the memory extension of the model.

We do not have a reference transcript available for the talks. Therefore, we manually labeled all detected occurrences of new words not present in the baseline hypothesis and all our approaches (4315) if they are correctly recognized or false positives.
Based on these labels, we calculate the forward transfer: For each new word $w$ that occurs somewhere in talk $i$ and $j$,
$i < j$,
we evaluate
if $w$ is recognized in talk $j$ using the model produced in iteration step $j-1$ (empty memory for MEM-CL).
We group according to how many training examples have been available until iteration step $j-1$ (false positives not counted), such that each bucket contains more than 300
samples.
Then, we calculate precision and recall for each bucket.
In total, the forward transfer is calculated on 402 unique words.

We see that for MEM-CL, the general performance (see Figure \ref{fig:cl}, left) is preserved, while for MEM-ALL and MEM-FOUND, the WER increases consistently over the course of the learning cycles (more than 12\,\% and 7\,\% relative respectively). This is because the MEM-ALL and MEM-FOUND models have to distinguish between more and more words in the memory which produces false positives. MEM-ALL and MEM-FOUND have 2199 and 692 memory entries at the end of the last learning cycle, respectively. For comparison, the baseline model
scores 4.39\,\% WER.

For forward transfer recall (see Figure \ref{fig:cl}, middle), we see that for MEM-ALL and MEM-FOUND the recognition performance of new words is at approximately 70\,\% after the new words are added to the memory. The models cannot utilize more occurrences of the same word. Therefore, performance does not increase over the course of the learning cycles. It even degrades a bit because the number of training samples per new word correlates with the total number of words in the memory, and with more words in the memory the performance degrades. In contrast, the performance of MEM-CL increases consistently to over 80\,\%
when more training samples of a new word arrive. 
For the forward transfer precision (see Figure \ref{fig:cl}, right) we find that MEM-CL has the best performance of all approaches and fewer false alarms are produced if more training samples per new word are available.
Words which MEM-CL recognized better than the baselines include backpropagate, CSPs, elementwise, frontend, LRU, MDP, MDPs, MRU, prefetcher, RNN, RNNs, SRAM, SIMD and tanh.

\subsection{Limitations}

The extraction of words from the slides
is not trivial. We took all words from the slides and filtered by the words in the training data.
Errors occur when the words on the slides are capitalized but in the transcript they should not be capitalized, and when the extracted words contain spelling errors.
A further problem arises when the morphological variances of known words are extracted, as their presence in memory increases the rate of false positives. Adding other morphological variances of the same word may help here. All these mentioned errors can propagate through the learning cycles. Moreover, training with pseudo-labels could lead to a degradation of the general performance. We did not observe that, however, one could restrict the training for the pseudo-labels to tokens that are part of
new words.
Other false positives can occur, e.g., when first a word like ISCA is learned followed by a word ISA which is written similarly. In this case, a small supervised intervention could help reduce the number of false positive utterances added to the adaptation data.


\section{Conclusion}

We demonstrated a self-supervised continuous learning approach for learning new words. This is done by iteratively extracting new words from slides of a given talk, detecting new words by a memory-enhanced ASR model, and using collected data for adapting low-rank matrix weights added to each weight matrix of the model. With this approach, we can increase the performance of new words as they occur more often to more than 80\,\% recall while the general performance of the model is preserved.

\section{Acknowledgements}


This research was supported in part by a grant from Zoom Video Communications, Inc.
Furthermore, the projects on which this research is based were funded by the Federal Ministry of Education and Research (BMBF) of Germany under the number 01EF1803B (RELATER),
the Horizon research and innovation program of the European Union under grant agreement No 101135798 (Meetween),
and the KIT Campus Transfer GmbH (KCT) staff in accordance with the collaboration with Carnegie – AI.
The authors gratefully acknowledge the support.
We thank Ngoc Quan Pham for providing his framework \mbox{NMTGMinor} to train our systems.


\bibliographystyle{IEEEtran}
\bibliography{mybib}

\end{document}